\documentclass{article}
\usepackage{spconf,amsmath,graphicx}
\usepackage{verbatim}
\usepackage{booktabs} 
\usepackage{caption}
\usepackage{xcolor}
\usepackage{pifont}
\usepackage{subcaption}
\usepackage{enumitem}
\usepackage{hyperref}
\usepackage{svg}
\usepackage{pmboxdraw}
\usepackage{comment}
\usepackage{algorithm}
\usepackage{algpseudocode}
\usepackage{amsmath}
\usepackage{array,etoolbox}
\preto\tabular{\setcounter{magicrownumbers}{0}}
\newcounter{magicrownumbers}
\def\rownumber{}

 \usepackage{soul}

\title{Towards Personalization of CTC Speech Recognition Models with Contextual Adapters and Adaptive Boosting}
\name{
\begin{tabular}{c}
Saket Dingliwal* \qquad  Monica Sunkara* \thanks{* Equal contribution} \qquad Sravan Bodapati \\ \textit{Srikanth Ronanki} 
\qquad  \textit{Jeff Farris}  \qquad \textit{Katrin Kirchhoff} 
\end{tabular}
}

\address{AWS AI Labs\\
\tt \{skdin, sunkaral\}@amazon.com}

\copyrightnotice{978-1-6654-7189-3/22/\$31.00~\copyright2023 IEEE}
\begin{document}
\ninept

\maketitle

\begin{abstract}
    End-to-end speech recognition models trained using joint Connectionist Temporal Classification (CTC)-Attention loss have gained popularity recently. In these models, a non-autoregressive CTC decoder is often used at inference time due to its speed and simplicity. However, such models are hard to personalize because of their conditional independence assumption that prevents output tokens from previous time steps to influence future predictions. To tackle this, we propose a novel two-way approach that first biases the encoder with attention over a predefined list of rare long-tail and out-of-vocabulary (OOV) words and then uses dynamic boosting and phone alignment network during decoding to further bias the subword predictions. We evaluate our approach on open-source VoxPopuli and in-house medical datasets to showcase a 60\% improvement in F1 score on domain-specific rare words over a strong CTC baseline.
\end{abstract}


\begin{figure*}[!t]
    \centering
    \includegraphics[width=0.7\linewidth]{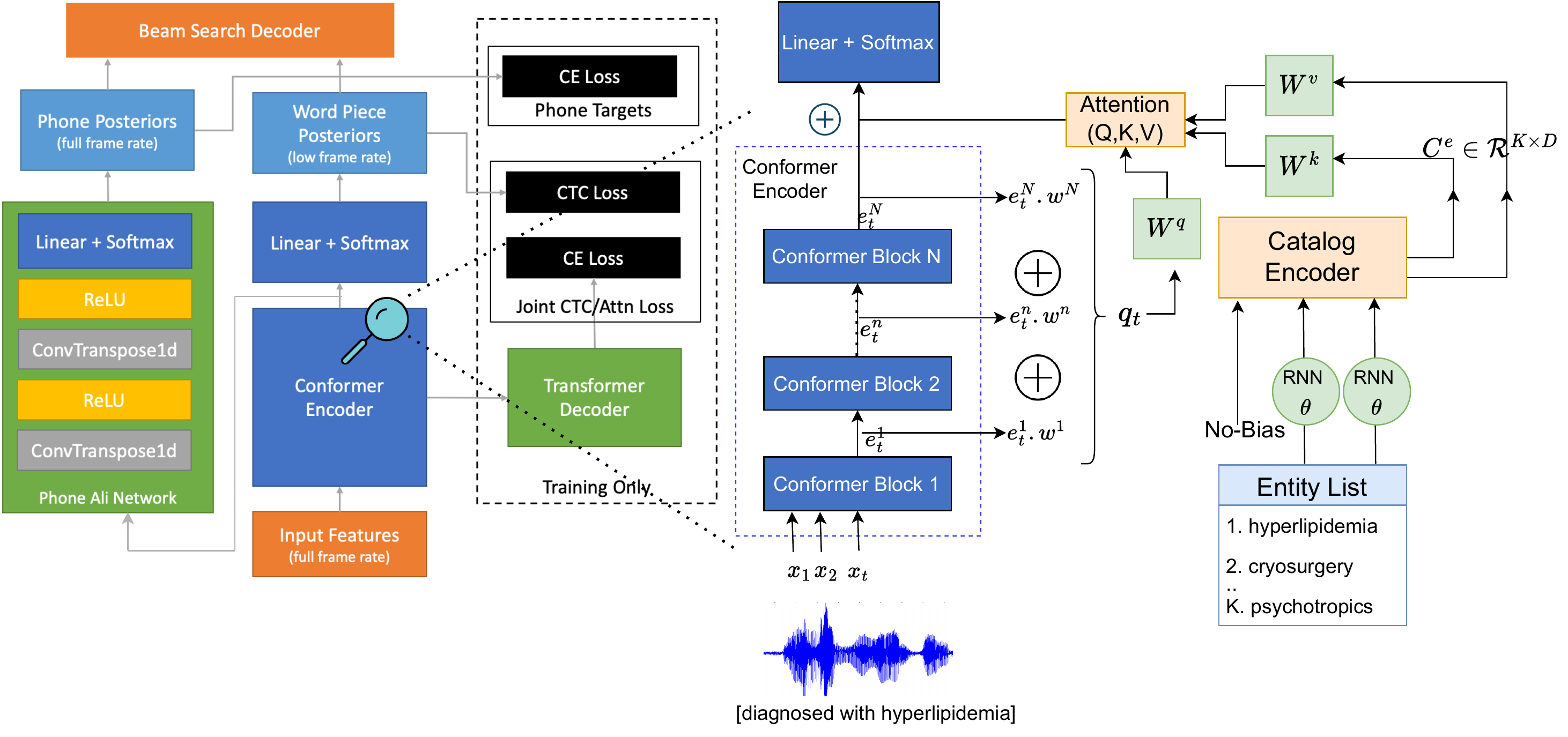}
    \caption{Multi-task encoder for Conformer CTC model (left) with  design for the Contextual Adapter (right) added to the encoder layers}
    \label{fig:ctc-attention}
    \vspace{-5mm}
\end{figure*}

\section{Introduction}
\label{sec:intro}

End-to-end (E2E) techniques such as CTC \cite{graves2013speech, salazar2019self}, Transducer \cite{graves2012sequence, he2019streaming} and Attention \cite{chan2015listen, lu2020exploring} have been successfully deployed for training large vocabulary speech recognition applications. Among all the techniques, single-pass CTC inference is an attractive choice for industry applications due to its low speed and memory footprint \cite{mcgraw2016personalized, laso, kim2020review, yao2021wenet}. Most recently, Conformer for speech recognition proposed in \cite{gulati2020conformer} combines Transformer and Convolution modules into ``Conformer'' blocks. This, in combination with CTC-Attention multi-task learning framework \cite{kim2017joint, watanabe2017hybrid} has become one of the popular recipes for training end-to-end speech recognition models \cite{guo2021recent}. While these models have achieved results on par with or better than conventional hybrid ASR systems \cite{li2020comparison, li2020developing, li2021recent}, they still struggle to recognize rare domain specific words and OOV words that are unseen during training \cite{sainath2018no, Bruguier2016LearningPP}. We therefore propose the biasing approaches for personalization of CTC models with a predefined list of uncommon but important entities which typically include contact names of users, names of clients for a company, medical terms or airport names.



\vspace{-2mm}
\subsection{Relation to prior work in personalization}
\label{ssec:related_work}

Several approaches have been proposed in the past for personalization, domain adaptation and contextual biasing of ASR systems. While some focused on adapting Language Models (LMs) in hybrid systems \cite{shenoy_1, ptuning, space-efficient}, others customized autoregressive architectures like Transducer and Attention-based encoder-decoder models \cite{jain2020contextual, le2021deep, gourav2021personalization, alexadapters, das2022listen}. We categorize these methods into (1) Early biasing, and (2) Decoder biasing. Methods in the former category focus on directly improving the likelihood of predicting custom words by training additional modules along with the base model 
while techniques in the latter tries to bias the decoding process with additional context as a post-processing step. Early biasing methods like training an attention module over custom list of user-specific words \cite{alexadapters, clas, phoebe}, that induce a better coverage of rare WordPieces have been shown to significantly benefit autoregressive models. In parallel, post-processing methods like shallow fusion \cite{shallowfusion}, deep fusion \cite{le2021deep}, leveraging grapheme-to-grapheme (G2G) \cite{le2020g2g} to produce additional pronunciation variants have also been shown to be useful for their simplicity. 

While some of these approaches are beneficial, accurate recognition of rare and OOV words still presents a challenge to CTC models because of two main reasons: First, CTC models can not influence future predictions based on previous output tokens, and hence cannot use the contextual predictions to customize beam paths in the early biasing methods. Second, the outputs of CTC include blanks and duplicate tokens which makes decoder biasing further complex. Additionally, the output units of CTC (or any other E2E) models are typically either graphemes or subwords, whose sequences do not correspond well to how they are pronounced \cite{le2021deep}. 

\vspace{-3mm}
\subsection{Novelty of this work}
\label{ssec:novelty}
In this work, we propose a novel two-phase approach which includes both early and decoder biasing techniques to address the challenges that are typical to personalization of CTC models. In the first phase, we use an attention-based Adapter module that uses speech representations from the encoder to attend over a list of custom entity words, thereby learning when to copy these entities to the beam path. In the second phase, we propose a combination of techniques to further boost these entities. Specifically, we propose on-the-fly dynamic boosting of subwords in combination with a prefix beam search decoder. In addition to this, we also adopt a phone alignment network and bias the subword predictions at inference time by utilizing both phone and subword output predictions. The use of phone predictions alleviates the problem of misrecognition of words whose pronunciations do not match a corresponding subword sequence. 

Overall, we showcase that our biasing approaches can significantly boost the performance of rare domain specific words and OOV words. Along with the improvements, the following characteristics make our joint model an important scientific and practical contribution. (1) Similar to \cite{clas, phoebe} and different from \cite{alexadapters}, our methods do not require any additional training data for personalization and the Adapter modules can be trained using the same data used for training the base model; (2) The parameters of the base model are not finetuned during personalization and hence the same model can be used on generic speech datasets with negligible effect; 
(3) Similar to \cite{clas, alexadapters}, the list of entities to the Adapter module can be changed at inference time as in a zero-shot setting and a common architecture fits all personalization use cases without the need for user or domain specific additional parameters; 
(4) We show that our joint approach can yield significant improvements regardless of the size of the base CTC model as well as the number of hours used for training; (5) We also provide analysis on the effect of biasing when the model is fed with a larger list of custom entities (more than 700). 
The main contributions of our work are summarized as follows:
\begin{itemize}[leftmargin=*,itemsep=1pt, topsep=2pt]
\item We propose an early biasing approach for Conformer-CTC architecture that leverages representations from multiple encoder layers to copy custom entities to the beam path. 
\item We also propose a combination of on-the-fly decoder biasing techniques for personalization of non-autoregressive CTC models during beam search decoding. 
\item We demonstrate how our proposed approaches work complementary to each other through extensive experimentation and ablation studies. On VoxPopuli dataset, our proposed joint model achieves 61.7\% and 31.4\% improvement in F1 scores of rare and OOV words respectively over the baseline CTC model.
\end{itemize}



\section{Model architecture}
\label{sec:background}

\subsection{Overview of Conformer CTC-Attention framework}
\label{ssec:ctc_overview}

The CTC-Attention framework \cite{kim2017joint}, can be broken down into three different components: \textit{Shared Encoder}, \textit{CTC Decoder} and \textit{Attention Decoder}. As shown in Figure \ref{fig:ctc-attention}, our \textit{Shared Encoder} consists of multiple Conformer \cite{gulati2020conformer} blocks with context spanning a full utterance. Each Conformer block consists of two feed-forward modules that sandwich the Multi-Headed Self-Attention module and the Convolution module \cite{gulati2020conformer}. A Convolution frontend with Conv2D layers downsamples the input features by a factor of four before feeding them to a Conformer Encoder. The \textit{CTC Decoder} consists of a linear layer with softmax activation, which transforms the \textit{Shared Encoder} output to the final output distribution over the WordPiece units. The \textit{Attention Decoder} consists of one Transformer decoder block and a softmax layer whose outputs are compared against reference outputs using a cross-entropy criterion. The intuition behind this joint modeling is to use a CTC objective function as an auxiliary task to train the \textit{Shared Encoder} within the multitask learning (MTL) framework whose loss is given below:
\vspace{-2mm}
\begin{equation}
\label{eq:alpha_ctc}
L_{MTL} = \alpha L_{CTC} + (1 - \alpha) L_{Att}    
\end{equation}

\noindent where $L_{MTL}$, $L_{CTC}$, and $L_{Att}$ represents multi-task learning loss, CTC-loss and \textit{Attention Decoder} cross-entropy loss respectively. $\alpha$ is a tunable parameter and represents the weight of CTC-loss whose value lies between 0 and 1. During inference time, only the \textit{Shared Conformer Encoder} and linear layer is used and the \textit{Attention Decoder} is discarded to ensure a low inference latency. Although \textit{Attention Decoder} can be used in the second pass to give a more accurate result, we did not perform any experiments in this direction and will be explored as part of future work. 

\vspace{-2mm}
\subsection{Phone alignment network with Conformer encoder}
\label{ssec:mtl_ctc}


Typically, the CTC decoder outputs (subword units and blank symbols) are collapsed and concatenated to form final word-level output. However, accurate recognition of rare and OOV words from subword units is a challenging problem and can be complemented with a phone prediction model. We use a supplementary context independent phone recognizer based on the work presented in \cite{zhao2021addressing}. During training, a joint phone alignment network as shown in Figure \ref{fig:ctc-attention} is trained from \textit{Shared Encoder} representations. This requires that we have phone alignments per frame derived from a hybrid system and a pronunciation lexicon. Since the Conformer architecture is low-frame-rate and the phone alignments are full-frame-rate, we use upsampling Convolutional layers (ConvTranspose1d) to transform the low-frame-rate encodings up to full-frame-rate. Per frame phone posteriors predicted from this network are optimized using a cross-entropy loss function and modified MTL loss is shown below:

\vspace{-5mm}
\begin{equation}
\label{eq:beta_ctc}
L_{MTL} = \alpha ((1-\beta) L_{CTC} + \beta L_{Ali}) + (1 - \alpha) L_{Att}
\end{equation}

\noindent where $L_{Ali}$ represents alignment loss and $\alpha$, $\beta$ are tunable parameters and represent the weights of  CTC-loss and alignment-loss respectively, whose value also lies between 0 and 1. If we already have a well trained CTC model, we could also directly add the phone recognizer and train it by freezing the \textit{Shared Encoder}. 
In Section \ref{ssec:mtl_biasing}, we present further details on biasing the subword predictions by utilizing the phone alignment network.


\section{Proposed Approach}
\label{sec:methods}


\vspace{-1mm}
\subsection{Early Encoder Biasing}
\label{ssec:encoder_biasing}

As shown in Figure \ref{fig:ctc-attention}, we propose to use Contextual Adapter to bias our pre-trained Conformer-CTC encoder. Similar to \cite{alexadapters}, our Contextual Adapter module consists of two components: Catalog Encoder and Biasing Adapter. The Catalog Encoder encodes a list of custom entities to produce corresponding embeddings while the Biasing Adapter uses output representations from the Conformer Encoder blocks to attend over these entities and learns to copy them to the output, when observed in the input speech. Figure \ref{fig:ctc-attention} summarizes the architecture we use for Contextual Adapters. Assume we have a set of $(K-1)$ custom entity words i.e. $\{w^j\}^{j=(K-1)}_{j=1}$, we tokenize each word with the same tokenizer as the base CTC model, to form a sequence of tokens $\{c_1^j, c_2^j ... c_n^j\}$. We then do an embedding lookup (parameterized by $\phi$) to create a sequence of vectors $\{c^j_{1:n}\}^\phi$.  We pass these sequences of subword embeddings through BiLSTM layers (parameterized by $\psi$) and use the final state as entity embedding i.e. $C^j$ as defined in equation \ref{eq:catalog}:

\vspace{-2mm}
\begin{equation}
\label{eq:catalog}
C^j= \textit{BiLSTM}_{\psi}(\{c^j_{1:n}\}^\phi) 
\end{equation}

We also learn an embedding vector $C^{\textit{nb}} \in \mathcal{R}^{D}$ for a special token $<\textit{no-bias}>$, so the model can attend to it when transcribing any word which is not present in the provided list of entities. Let $C^e \in \mathcal{R}^{K \times D}$ represent concatenated embeddings of dimension $D$ of the entity items plus the no-bias token. 

The Biasing Adapter then uses these embeddings as keys and values in an attention module to obtain a context vector that is added to the output of the last layer of the Conformer Encoder. More formally, let $\{e^i_t\}^{i=N}_{i=1}$ represent the output of each of the N Conformer blocks present in the Encoder at time step $t$. We define a set of $N$ trainable weights $\{w^i\}^{i=N}_{i=1}$ to create our attention query $q_t$ as described in Equation \ref{eq:adapter} and the output of the Biasing Adapter can be defined as $b_t$: 
\vspace{-4mm}
\begin{equation}
\label{eq:adapter}
b_t = \textit{Attn}(W^q q_t, W^k C^e, W^v C^e) \text{\;where\;}  q_t = \sum_1^N w^i e^i_t
\vspace{-4mm}
\end{equation}

\noindent where $W^q, W^k, W^v$ are query, key and value weight matrices respectively and $\textit{Attn}(.)$ is an attention function as defined in \cite{attention}. The output of Biasing Adapter is added to final encoder output before passing through a softmax layer
\vspace{-1mm}
\begin{equation}
    e_t^N = e_t^N + b_t 
\end{equation}
\vspace{-5mm}

The formulation of our proposed Contextual Adapters is adapted from \cite{alexadapters}, and modified for a non auto-regressive CTC model as follows. It leverages information available in different encoder layers, rather than just the last layer. This is essential specially for CTC architectures like ours where the last encoder layer representations are closer to output representations (word-pieces) as compared to other architectures that rely on multiple decoder layers. Also, note that for CTC models, the learnt catalog embeddings ($C^e$) are not specific to the content/semantics of the entity but are just a representation of the sub-word sequence of the entity. Different from \cite{alexadapters}, we do not use any additional data for training the adapters in our experiments. We introduce a simple trick to reuse the generic data used for training the base model to train our version of Contextual Adapters. From the training transcripts, we identify a list of rare words (any word that appears less than 13 times) and the corresponding utterances. Then for each of these utterance, the list of input custom entity words simply comprises of a random subset from all the rare words and the rare word present in that utterance. This helps train the Biasing Adapter such that it uses the speech signal to understand if it needs to copy the word-piece sequence of the rare word to the output or choose the no-bias for any time frame.  This training method ensures that our personalization approaches are truly zero-shot and no additional labeled data is required for adopting them in practice. Formally, let $\gamma$ represent all the parameters of the base Conformer CTC model obtained after minimizing loss $L_{MTL}$ as defined in \ref{eq:alpha_ctc} on the dataset $\mathcal{D}$. Then we jointly optimize all the parameters in the Contextual Adapters using the subset of the dataset $\mathcal{D}^{\textit{rare}} \subset \mathcal{D}$ as defined above, keeping $\gamma$ fixed as defined in Eqn \ref{eq:adaploss} below
\vspace{-4mm}
\begin{equation}
    \label{eq:adaploss}
    \{\psi, \phi, C^{\textit{nb}}, \{w^i\}^{i=N}_{i=1}, W^q, W^k, W^v \} = \text{argmin} \sum^{\mathcal{D}^{\textit{rare}}} L_{MTL}(\gamma, \cdot)
\vspace{-1mm}
\end{equation}

At inference time, we swap in a fixed list of custom entity words for all the utterances in the test dataset. Since the catalog of entities is fixed, we create the entity embeddings $C^e$ only once and save it in cache. This ensures we do not add any observable change to the inference latency of the system. Further, if for any frame, the attention weight given to the no-bias term is the maximum among all other catalog entities, we explicitly set $b_t = 0$ so that there is no regression/change in performance of the system on generic dataset/speech segments without the entity words.

\vspace{-2mm}
\subsection{Decoder Biasing Methods}
\label{ssec:decoderbiasing}

\subsubsection{Adaptive subword boosting in beam search decoding}
\label{ssec:adaptive_boosting}

The process of tuning boosting probabilities may lead to inconsistent results and is considered a sub-optimal approach. Therefore, we implemented an adaptive on-the-fly boosting of the vocabulary of interest where biasing is done at subword level during beam search decoding before the pruning stage. As described in \cite{liptchinsky2017letter}, our beam search decoder performs a simple beam-search with beam thresholding, histogram pruning, and language model smearing. The decoder accepts unnormalized acoustic scores ($g_{i,j}(.)$) as input and attempts to maximize the following:
\vspace{-3mm}
\begin{equation}
\begin{aligned}[b]
& L(\theta ) = \underset{\pi \in G_{lex} (\theta , T)}{logadd} \sum_{t=1}^{T} (f_{\pi_t}(x) + g_{\pi_{t-1},\pi_t}(x) + h_{\pi_t}(x)) \\ 
& + \alpha logP_{lm}(\theta ) + \beta |\{i|\pi _{i} = \# \}|,
\end{aligned}
\end{equation}
\noindent where the “$logadd$” operation (also called “log-sum-exp”) is defined as $logadd(a, b) = log(exp(a) + exp(b))$. $G_{lex}(\theta, T)$ is a graph constrained by base lexicon over $T$ frames for a given transcription $\theta$, and $\pi=\pi_{1},...,\pi_{t} \in G_{lex}(\theta,T)$ is a path in this graph representing a valid subword sequence for this transcription. At each time step $t$, $f_{\pi_{t}}(x)$ denotes the log-probability by the CTC model (given an acoustic sequence $x$) and $h_{\pi_{t}}(x)$ denotes the additional boosting score. $P_{lm}(\theta)$ is the probability of the language model given a transcription $\theta, \alpha$ and $\beta,$ are hyper-parameters which control the weight of the language model, and silence (\#) insertion penalty, respectively.

Let $\hat{y}(t,k)$ denote the log-probability distribution by the CTC model in sorted order at time $t$, with k representing the subword candidate index in the top-K list. To compute the boosting score, we use a \textit{trie} ($T_{vocab}$) constructed from the list of custom entities and boost the rest of the sequence if the first subword (trie node) occurs in the top-K list. The boosting score for each subword candidate at time step $t$ is determined dynamically by its difference in log-probability with the top-1 hypothesis. 
\vspace{-2mm}
\begin{equation}
\label{eq:delta_ctc}
h_{\pi_t}(x) = \left\{\begin{matrix}
 \delta * (\hat{y}(t,k)-\hat{y}(t,1)) & \pi_{t} \in T_{vocab} \\ 
 0 & Otherwise \\
\end{matrix}\right. 
\end{equation}
\noindent where $\delta$ denotes the boosting scale. To reduce false positives, we follow an inverse \textit{sigmoid} function for computing the boosting scale as shown below:
\vspace{-2mm}
\begin{equation}
\label{eq:gamma_ctc}
\delta = 1 / (1 + e^{(\hat{y}(t,k)-\hat{y}(t,1)- 0.5*k)/0.1*k})
\vspace{-3mm}
\end{equation}
\vspace{-3mm}

\vspace{-1mm}

\noindent \textbf{Unigram boosting:} We also consider modifications to the external N-gram LM before performing fusion with a beam search decoder. In theory, one simple way of biasing is to add OOV words to the external LM with a fixed high unigram probability. This can also be achieved by adding an OOV class (instead of adding all words) in order to keep the LM unmodified during inference. In this work, we add not only OOV words, but also other common and rare words that are tested as part of experiments presented in the Section \ref{sec:experiments}.

\subsubsection{Phone alignment network for biasing}
\label{ssec:mtl_biasing}

In this section, we describe two biasing techniques using the per-frame phone predictions generated from the phone alignment network described in the Section \ref{ssec:mtl_ctc}.

\vspace{2mm}
\noindent \textbf{Phonetic distance based rescoring:}
Although unigram boosting helps to a certain extent, rare and OOV words are still considered unknown tokens in higher order N-grams. In our preliminary experiments, we observed that the words that are boosted during on-the-fly rescoring are less preferred due to shallow fusion with an external language model. We therefore propose to use phonetic distance based rescoring of the N-best lists to bias those hypotheses that contain rare and OOV words. Let $X = \{x_1, x_2,...x_T\}$ be a sequence of per-frame phone posteriors generated from the phone alignment network. Let $Y^n = \{y^n_1, y^n_2,...y^n_{L_n}\}$ be a phone sequence corresponding to the $n^{th}$-best hypothesis (denoted by $Y_w^n$) from the N-best lists generated from the \textit{CTC decoder} by performing shallow fusion with an external language model. We then perform forced alignment of each hypothesized phone sequence from the N-best list to the per-frame phone posteriors generated from the phone alignment network using a dynamic time warping (DTW) algorithm. The $X$ and $Y^n$ sequences can be arranged to form a $T$-by-$L^n$ grid, where each point (i, j) is the alignment between $x_i$ and $y^n_j$. A warping path $W$ maps the elements of $X$ and $Y^n$ to minimize the distance between them. The optimal path to ($i_k$, $j_k$) can be computed by:
\begin{equation}
    {D_{min}(i_k,j_k) = \underset{i_{k-1},j_{k-1}}{min} D_{min}(i_{k-1},j_{k-1})+d(i_k,j_k|i_{k-1},j_{k-1})}
\end{equation}

\noindent where d is the Euclidean distance. Let us define $d^n$ as the DTW cost incurred to align $j^{th}$ feature of $Y^n$ with the $i^{th}$ feature vector of $X$ and the overall path cost ($D^n$) can be calculated as:
\begin{equation}
    D^n = \sum_k d^n(i_k, j_k)
\end{equation}
\vspace{-4mm}

\noindent where $i_k=T$ and $j_k=L_n$. We insert a \textit{/silence/} phone between each word to allow for pauses between words during the forced alignment. We also use a fixed scale of $0.1$ for DTW cost and the overall cost defined below is used for rescoring the N-best lists:
\begin{equation}
    rescore = P_{CTC}(Y_w^n) + 0.1 * D^n
\end{equation}


\vspace{2mm}
\noindent \textbf{Use of pronunciation lexicon:} Traditional hybrid systems use phonetic lexicons and/or clustered context-dependent acoustic targets to recognize long-tail words. Similarly, we hypothesize the phone predictions to be more accurate than subword predictions for rare and OOV words. In this approach, we take the 1-best hypothesis from the \textit{CTC Decoder} after performing phonetic distance based rescoring and tokenize it into words. The word boundaries generated during forced alignment are used to retrieve corresponding per-frame phone predictions. We then perform window-based smoothing to replace any spurious predictions and collapse consecutive repetitive predictions into a single phone. The pronunciation obtained in this way is compared to the lexicon-derived pronunciation of each vocabulary from the user provided biasing list. If an exact match is found, we replace the word in the 1-best hypothesis with matched vocabulary. Although this is a heuristic based approach, it has been shown to generalize across datasets with no additional tuning. 

\begin{table*}[t]
\caption{Performance comparison of personalization methods on VoxPopuli dataset}
\centering
\begin{tabular}{|@{\makebox[2em][r]{\rownumber\space}}|l|c|c|c|c|c|c|c|}
\toprule

Model                                                              & \textbf{WERR}          & \multicolumn{3}{c|}{\textbf{Rare}}                                           & \multicolumn{3}{c|}{\textbf{OOVs}}                  \\
\gdef\rownumber{\stepcounter{magicrownumbers}\arabic{magicrownumbers}} 
                                                                    &              (\%age)        & F1                   & P            & R              & F1                   & P  & R  \\
\midrule

Baseline Conformer-CTC (Large)                       &        -          &            49.4          &      97.7                &       33.1               &           59.0          &  100.0       &    41.8 \\
\midrule
Row 1 + Contextual Adapter (top layer)              &          -1          &   74.4              &         86.5            &               65.4       &              68.5        &     69.8      & 67.3       \\
\hspace{2.5em}  \hspace{0.5em} + Enforce no-bias token       &  -0.5          &        71.9              &          86.7            &       61.4               &  68.0                    &         70.6             &      65.5             \\
\hspace{3.0em}  \hspace{0.5em} + Intermediate Layers               &         2           &    77.0                  &                95.3  &     64.6                &       69.3               &    76.1      &     63.6  \\
\midrule
Row 1 + Adaptive subword boosting                                    &        -1          &          72.5            &            87.4         &      62                &    68.2                  &     79.5     &  59.8       \\
\hspace{2.5 em}  \hspace{0.5em} + Phonetic distance rescoring    & -0.8  &       74.6                               &           89.5           &    63.9                  &    69.1      &  80.1   & 60.7  \\

\hspace{3.0em}  \hspace{0.5em} + Pronunciation lexicon   &     0                 &           76.2           &           89.1          &    66.5                  &     71.6     &  78.5   &  65.7 \\
\hspace{3.5em}  \hspace{0.5em} + G2G       &               0.2     &          76.7            &   88.6                   &         67.6             &     72.5     &  78.1     & 67.6 \\
\hspace{4.0em}  \hspace{0.5em} + Shallow fusion with boosted LM   &   0.4    &  77.4                                 &             90.1         &         67.9             &      73.2    &   78.6   & 68.5 \\
\midrule
Proposed joint model (Row 4 + Row 9)                                                      &         1.2            &         79.9             &        92.6             &            70.2        &              77.5        &      81    &    74.3   \\
\toprule
\end{tabular}
\label{tab:voxpopuli_62k}
\vspace{-5mm}
\end{table*}

\vspace{-2mm}
\subsubsection{Leveraging phoneme similarity}
\label{ssec:g2g}

Previous work on personalization for RNN-T models \cite{le2021deep} leveraged G2G to generate additional pronunciation variants for domain-specific entities during decoding. Similarly, we generate few variants using phoneme similarity and decompose them into subwords while maintaining the word-level label. With this approach, the probability of predicting the actual word slightly improves with an increase in the number of unique G2G variants.

\vspace{-1mm}
\section{Experiments}
\label{sec:experiments}

\vspace{-2mm}
\subsection{Data}
\label{ssec:data}


\noindent \textbf{Training data}: We train our base Conformer CTC model on two different corpora in size i.e., a large 50k+ hour English corpus and a smaller subset of 5k hour corpus, sampled from in-house paired audio and text data. These two data regimes are representative of a wide range of end-to-end ASR systems for various different speech applications. Both corpora include audio files with a good mix of accents, speakers, sampling rates and background noise. 


\vspace{2mm}
\noindent \textbf{VoxPopuli}: For evaluation, we use a 5 hours test subset of publicly available VoxPopuli English data \cite{voxpopuli} with an average recording length of 9.6 seconds. We extract out entities from the reference text based on the frequency of each token in our larger training corpus. Any token that appears in the reference text but occurs less than 50 times in the training transcripts is called rare while OOVs are the words which are unseen during training. Our evaluation subset of the dataset contains 44407 tokens in total, of which 574 (96 unique) are rare and 55 (42 unique) are OOVs. 

\vspace{2mm}
\noindent \textbf{Medical}: We also evaluated on an in-house medical dataset consisting of 609 recordings with each audio spanning an average of 147 seconds (25 hours in total). For preparation of the dataset, speakers of different genders, accents and ages read written conversational style utterances in different acoustic settings (background noise). These utterances contain a desired list of 755 medical entities  (716 rare and 39 OOVs) like names of medicines and diseases eg. ‘acetabulofemoral’, ‘ecchymoses’, etc which are long (average  9.2 characters per word) and hard to be recognized by any generic ASR system and hence used to gauge the performance of our proposed methods. This dataset contains a total of 189380 tokens, of which 3625 are rare tokens and 198 OOV tokens. We use this dataset to demonstrate the efficacy of our proposed methods while scaling to a larger list of medical terms. For experimentation, we also trained a medical domain-specific language model on an internal medical text corpus with 3.9 million words (21k unique tokens).

\vspace{-2mm}
\subsection{Experimental Setting}
\label{ssec:baseline}
\noindent \textbf{Baseline}: We train two different mixed-bandwidth 16k models of the multi-task Conformer CTC-Attention architecture as described in Section \ref{ssec:mtl_ctc}. For the large model, we stacked 20 layers of Conformer blocks with 8 attention heads and roughly 142.5 million parameters and trained on our large English corpus, while for the small model, we stacked 16 blocks with 4 attention heads and 30.2 million parameters and trained on 5k hour subset. We train a sentence-piece tokenizer with token size of 2048 and 1024 for large and small model respectively. Both baselines are trained with ADAM optimizer with a learning rate of 0.001. Further, we train an external 4-gram language model (LM) for shallow fusion. During inference, we use a beam size of 50 and LM weight of 0.6 in all our experiments. All our work is implemented in the open-source ESPnet tool \cite{watanabe2018espnet}. 


\vspace{1mm}
\noindent \textbf{Encoder-biasing}: 
We train the Contextual Adapters on the same subset of data used for training the generic CTC model. To save compute costs, we only use the 10k hour subset from our large corpus for training the adapters. For both the models, $D = 128$ (catalog embedding dimension) and we vary $k$ (catalog size) from 30 to 250 during training (starting with 30 and increasing it by 4 every epoch to a maximum of 250). 
We only use 1 attention head for the Biasing Adapter and attention dimension of 128. We freeze all other parameters of the base model as we update the Adapter parameters (1.32 million, which is $<1\%$ of the parameters of the base large model). We train two versions of the Adapters for the larger model, one where attention query $q_t$ is with weighted sum of 6, 12 and 20th encoder layer (called 'intermediate layers') and the other with just the last layer \cite{alexadapters} (called 'top layer'). Further, we run inference with and without enforcing $b_t = 0$ for frames with maximum weight to no-bias token (denoted as 'enforce no-bias'). We use the best version for the 5k-hr model ablation.  

\vspace{1mm}
\noindent \textbf{Decoder-Biasing}: At each step of beam search decoding we consider a top 10 candidates for expanding our decoding tree. Since the BPE model typically generates a unique segmentation of words, subwords into which rare words are split have poor probability. However, we noticed that often a different subword sequence which corresponds to a rare word occurs in the beam paths. To overcome this issue, we can use a dropout in the BPE model to generate multiple variations of segmentation of rare words. In our experiments, we generated up to 10 variations of segmentation for the rare words that we boost in prefix based adaptive subword boosting. We use a unigram boosting score of -0.2 in our domain language model after careful tuning on the dev dataset. 

\vspace{1mm}
\noindent \textbf{Joint model}: When combining encoder and decoder biasing methods, we do not tune any hyper-parameters of the individual techniques again. All the results presented in this paper are obtained by direct combination, without changing the individual methods.

\vspace{-2mm}
\section{Results}
\label{sec:results}
\vspace{-2mm}
Tables \ref{tab:voxpopuli_62k} and \ref{tab:voxpopuli_5k} present the results of our proposed personalization methods when biased with rare and OOV words found in the VoxPopuli test dataset. In Table \ref{tab:voxpopuli_62k}, we compare both encoder and decoder biasing methods as well as their joint combination against a strong Conformer-CTC baseline. All the methods used an external 4-gram language model in fusion with a beam search decoder. The numbers reported in the first column are Word Error Rate Reduction \%age (WERR) over the baseline 9.1 WER. Overall, the proposed approaches outperform the baseline in both WER and F1 score. 

\vspace{1mm}
\noindent \textbf{Encoder-biasing}: In the encoder biasing, the addition of the Contextual Adapter to the top layer alone significantly improved the F1 score of both rare words and OOVs, but also resulted in a small increase in WER. This is likely due to the high attention weight from the biasing adapter forcing the encoder to copy a word and not allowing the model to recover from any false positives. However, when adapters are introduced in intermediate layers with learnable weights and the no-bias token is enforced, the overall WER is improved over the baseline model along with F1 score. 

\vspace{1mm}
\noindent \textbf{Decoder-biasing}: In decoder biasing, we applied a combination of methods on top of the baseline beam search decoding to recover domain-specific vocabulary as summarized in the Table \ref{tab:voxpopuli_62k}. First we implemented a subword level boosting on all candidates in the beam paths before pruning as described in Section \ref{ssec:adaptive_boosting} which led to significant improvements across all categories of vocabulary using this technique. We noticed that even though the subword boosting technique was very effective, the correct hypothesis with domain-specific vocabulary doesn’t surface in the 1-best but remains in the N-best which accounted for 10\% of errors. To alleviate this, we applied phonetic distance based rescoring which led to a relative improvement of ~3\% in rare vocabulary recognition. However, we only noticed a ~1.3\% improvement in OOV. This is probably due to the model assigning low scores to subwords that make up the OOV words, resulting in their absence in the N-best hypotheses. We then applied a pronunciation lexicon to further improve the 1-best hypothesis. As hypothesized, the phone predictions of the model were more accurate than subword predictions in certain cases, thereby leading to improvements in both rare and OOV words. Finally, we applied G2G and a simple technique of shallow fusion with a boosted language model as described in \ref{ssec:decoderbiasing} which improved performance across both rare and OOV words due to boosting at word level.

\begin{table}[t]
\footnotesize
\centering
\caption{Performance comparison with smaller subset of training data (5000 hours) on VoxPopuli dataset}
\vspace{-3mm}
\begin{tabular}{l|c|c|c}
\toprule
\textbf{Model} & \textbf{WERR}  & \textbf{F1} & \textbf{F1} \\
& \%age & Rare & OOV \\
\midrule
 Baseline Conformer-CTC (Small) & - & 32.9  & 47.4 \\
\midrule
Encoder Biasing (Row 4 from Table \ref{tab:voxpopuli_62k})  & 0.6 & 53.6 & 64.4 \\
Decoder Biasing (Row 9 from Table \ref{tab:voxpopuli_62k}) & 0.9 & 54.2 & 67.5 \\
Proposed Joint Model  & 1.6 & 56.1 & 69.8 \\
\bottomrule
\end{tabular}
\label{tab:voxpopuli_5k}
\vspace{-5mm}
\end{table}

\vspace{1mm}
\noindent \textbf{Joint model}: When we experimented with the combined model containing both the encoder and decoder biasing methods, we observe a significant improvement in F1 scores. From our error analysis, we noticed that the main challenge for poor recognition of rare and OOV words is either beam paths containing rare subword sequences getting pruned due to low probability or the rare subwords not being present in the top-k candidates. Our joint model is very effective in addressing these issues due to the complimentary nature of the proposed approaches. The encoder biasing method helps very effectively in generating higher probability scores of the rare subwords it copies which helps in preventing the rare subwords from getting pruned in the beam path or top-k candidates. This is further improved by applying the decoder biasing techniques in boosting the rare word candidate paths to 1-best which led to improvements in F1 across both rare and OOV words. We also observed that the Contextual Adapter biasing approach performs better than only subword boosting which gave the most gains in decoder biasing methods across both categories. However, when combined with other approaches such as phonetic distance based rescoring, pronunciation based lexicon lookup and G2G, decoder biasing has better recall on rare and OOVs. We hypothesize that encoder-biasing method tends to mis-recognize words whose pronunciations differ from the spellings as it does not have access to phonetic information of the entity word. However, the proposed decoder-biasing methods leverage phone sequences from phone decoder and lexicon based pronunciation of the word to boost or correct the rare, OOV words in the n-best list and the final 1-best. Decoder based biasing methods also provide the flexibility to not rely on the underlying model performance by providing several knobs to finetune for boosting weights and nbest list correction methods.  
Table \ref{tab:qual_results}, summarizes the qualitative examples of the joint model outputs. 



\begin{table}[t]
\footnotesize
\caption{Performance comparision of proposed methods with larger catalog size (755) on medical dataset}
\vspace{-3mm}
\begin{tabular}{l|c|c|c}
\toprule
\textbf{Model}  & \textbf{WERR}  & \textbf{F1} & \textbf{F1} \\
& \%age & Rare & OOV \\
\midrule
 Baseline Conformer-CTC (Large) &   - & 36.4 & 15.7 \\
 \hspace{0.5em} + Shallow Fusion (SF) with medical LM &   11.5 & 39.2 & 15.4 \\
 \midrule
SF + Encoder Biasing &   13.2 & 58.1 & 29.8 \\
SF + Decoder Biasing   & 4.9 & 57.9 & 22.4 \\ 
SF + Proposed Joint Model & 5.6  & 62.1    & 27.2 \\
\bottomrule
\end{tabular}
\label{tab:medical_results}
\end{table}

\begin{table}[t]
\footnotesize
\caption{Example utterances from VoxPopuli dataset with outputs from our proposed joint model}
       \vspace{-4mm}
        \begin{tabular}{p{8.3cm}}
            \toprule
            \textbf{baseline:} \textit{We will be able to start our \textcolor{red}{\hl{trio}} negotiations} \\
            \textbf{joint model:} \textit{We will be able to start our \textcolor{blue}{\hl{trialogue}} negotiations} \\
            \midrule
            \textbf{baseline:} \textit{Resignation of professor \textcolor{red}{\hl{galeo}} as chairman of the snc} \\
             \textbf{joint model:} \textit{Resignation of professor \textcolor{blue}{\hl{ghalioun}} as chairman of the snc}\\
             \midrule
             \textbf{baseline:} \textit{president \textcolor{red}{\hl{taani}} has stated that industry is at the heart of europe} \\
             \textbf{joint model:} \textit{president \textcolor{blue}{\hl{tajani}} has stated that industry is at the heart of europe} \\
             \midrule
             \textbf{baseline:} \textit{Mediation of pat cox and alexander \textcolor{red}{\hl{KSSK}}} \\
             \textbf{joint model:} \textit{Mediation of pat cox and alexander \textcolor{blue}{\hl{KWANIEWSKI}}} \\
             \midrule
             \textbf{baseline:} \textit{How is it possible for jan  \textcolor{red}{\hl{marle}} to completely disappear} \\
             \textbf{joint model:} \textit{How is it possible for jan \textcolor{blue}{\hl{marsalek}} to completely disappear} \\
             \midrule
             \textbf{baseline:} \textit{Fellow \textcolor{red}{\hl{creations}} and other nations} \\
             \textbf{joint model:} \textit{Fellow \textcolor{blue}{\hl{croatians}} and other nations} \\
             \midrule
             \textbf{baseline:} \textit{A significant number of \textcolor{red}{\hl{tanous}} national minorities} \\
             \textbf{joint model:} \textit{A significant number of \textcolor{blue}{\hl{autochthonous}} national minorities} \\
            \bottomrule
        \end{tabular}
    \label{tab:qual_results}
     \vspace{-7mm}
\end{table}

\vspace{-3mm}
\subsection{Ablation study with model trained on smaller corpus}
We evaluated our model performance when trained on a smaller corpus of 5k hour. Table \ref{tab:voxpopuli_5k} summarizes the results comparing different proposed approaches against the baseline (with 12.5 WER). We observe that both methods improved WER and F1 of rare and OOV entities with decoder biasing methods performing better on OOV words. This is because the decoder biasing methods such as lexicon lookup and G2G work effectively even if the subword sequences related to OOV are absent in the top candidates of the decoding. Our joint proposed model has the best performance with 70.5\% and 47.3\% relative improvements in F1 of rare and OOV words respectively over the baseline.


\begin{table}[t]
\footnotesize
\centering
\caption{Performance comparison of our methods customized for VoxPopuli entity list on general test set }
\begin{tabular}{l|c}
\toprule
\textbf{Model} & \textbf{WERR (\%age)}  \\
\midrule
 Baseline Conformer-CTC (Large)  & - \\
\midrule
Encoder Biasing & -0.34 \\
Decoder Biasing & -0.39 \\
Proposed joint model  & -0.42  \\
\bottomrule
\end{tabular}
\label{tab:generaltest}
\vspace{-4mm}
\end{table}

\vspace{-3mm}
\subsection{Effect of personalization on general dataset}
Personalization or adaptation of models to a specific use case or domain often leads to degradation in performance on general datasets. To demonstrate that our proposed approaches are generic, we evaluated the different methods personalized for VoxPopuli dataset on a general test set not containing entities (comprising of 1671 audio files) sampled from an original VoxPopuli data distribution with a baseline 8.8 WER. As summarized in table \ref{tab:generaltest}, we noticed a very small degradation with individual encoder and decoder biasing approaches. The joint model had a slightly higher degradation of 0.42\% than individual approaches due to overbiasing. However, this degradation (less than 0.5\%) is negligible compared to the significant improvements of rare and OOV words on the overall dataset.


\vspace{-3mm}
\subsection{Efficacy of proposed approaches on large catalog size}

Table \ref{tab:medical_results} presents the results on a large catalog size typically found in the medical domain. We find that our Contextual Adapter method scales very well to a larger custom list. We observed a significant 48\% relative improvement in F1 score on rare words along with a small improvement in WER. Although, the Adapters were exposed to a maximum of 250-sized entity list while training, they were observed to be performing well for a list that is roughly three times in size. However, we see a regression in WER as compared to a simple shallow fusion approach on using decoder-biasing methods on such a large list. This is not surprising because if we boost a large number of words while decoding, it is likely that many similar sounding words will match, thereby causing a drop in precision. On the other hand, an encoder-biasing method like Adapters rely on the speech signal to selectively choose when to copy a word, rather than simply copying it every time. Particularly, in our technique 'enforce no bias', we explicitly ensure that system output does not change for the frames where no-bias token has the maximum attention weight. This analysis can be useful for the research community to choose among various proposed personalization methods based on their use case, size of custom entities and trade-offs.

\vspace{-3mm}
\section{Conclusions}
\label{sec:conclusions}

In this paper, we proposed a combination of techniques to bias a conformer-CTC model to significantly improve the recognition of domain specific vocabulary, especially rare (long-tail) and OOV words. We explored both encoder biasing and on-the-fly decoder biasing techniques and compared their performance under two different data settings with base models trained on both large and small corpora. We demonstrated the efficacy of our proposed approaches with significant improvements in F1 score when biased with a large catalog. Our proposed model showed significant improvements of 60.3\% and 30\% in relative F1 score on rare and OOV words respectively on the VoxPopuli dataset with negligible degradation (less than 0.5\%) in WER on general utterances. 

\bibliographystyle{IEEEbib}
\bibliography{refs}

\begin{thebibliography}{10}

\bibitem{graves2013speech}
Alex Graves, Abdel-rahman Mohamed, and Geoffrey Hinton,
\newblock ``Speech recognition with deep recurrent neural networks,''
\newblock in {\em Proc. ICASSP}, 2013, pp. 6645--6649.

\bibitem{salazar2019self}
Julian Salazar, Katrin Kirchhoff, and Zhiheng Huang,
\newblock ``Self-attention networks for connectionist temporal classification
  in speech recognition,''
\newblock in {\em Proc. ICASSP}, 2019, pp. 7115--7119.

\bibitem{graves2012sequence}
Alex Graves,
\newblock ``Sequence transduction with recurrent neural networks,''
\newblock {\em arXiv preprint arXiv:1211.3711}, 2012.

\bibitem{he2019streaming}
Yanzhang He, Tara~N Sainath, Rohit Prabhavalkar, Ian McGraw, Raziel Alvarez,
  Ding Zhao, David Rybach, Anjuli Kannan, Yonghui Wu, Ruoming Pang, et~al.,
\newblock ``Streaming end-to-end speech recognition for mobile devices,''
\newblock in {\em Proc. ICASSP}, 2019, pp. 6381--6385.

\bibitem{chan2015listen}
William Chan, Navdeep Jaitly, Quoc~V Le, and Oriol Vinyals,
\newblock ``Listen, attend and spell,''
\newblock {\em arXiv preprint arXiv:1508.01211}, 2015.

\bibitem{lu2020exploring}
Liang Lu, Changliang Liu, Jinyu Li, and Yifan Gong,
\newblock ``Exploring transformers for large-scale speech recognition,''
\newblock {\em arXiv preprint arXiv:2005.09684}, 2020.

\bibitem{mcgraw2016personalized}
Ian McGraw, Rohit Prabhavalkar, Raziel Alvarez, Montse~Gonzalez Arenas,
  Kanishka Rao, David Rybach, Ouais Alsharif, Ha{\c{s}}im Sak, Alexander
  Gruenstein, Fran{\c{c}}oise Beaufays, et~al.,
\newblock ``Personalized speech recognition on mobile devices,''
\newblock in {\em Proc. ICASSP}, 2016, pp. 5955--5959.

\bibitem{laso}
Ye~Bai, Jiangyan Yi, Jianhua Tao, Zhengkun Tian, Zhengqi Wen, and Shuai Zhang,
\newblock ``Listen attentively, and spell once: Whole sentence generation via a
  non-autoregressive architecture for low-latency speech recognition,''
\newblock {\em arXiv preprint arXiv:2005.04862}, 2020.

\bibitem{kim2020review}
Chanwoo Kim, Dhananjaya Gowda, Dongsoo Lee, Jiyeon Kim, Ankur Kumar, Sungsoo
  Kim, Abhinav Garg, and Changwoo Han,
\newblock ``A review of on-device fully neural end-to-end automatic speech
  recognition algorithms,''
\newblock in {\em Asilomar Conference on Signals, Systems, and Computers},
  2020, pp. 277--283.

\bibitem{yao2021wenet}
Zhuoyuan Yao, Di~Wu, Xiong Wang, Binbin Zhang, Fan Yu, Chao Yang, Zhendong
  Peng, Xiaoyu Chen, Lei Xie, and Xin Lei,
\newblock ``Wenet: Production oriented streaming and non-streaming end-to-end
  speech recognition toolkit,''
\newblock {\em arXiv preprint arXiv:2102.01547}, 2021.

\bibitem{gulati2020conformer}
Anmol Gulati, James Qin, Chung-Cheng Chiu, Niki Parmar, Yu~Zhang, Jiahui Yu,
  Wei Han, Shibo Wang, Zhengdong Zhang, Yonghui Wu, et~al.,
\newblock ``Conformer: Convolution-augmented transformer for speech
  recognition,''
\newblock {\em arXiv preprint arXiv:2005.08100}, 2020.

\bibitem{kim2017joint}
Suyoun Kim, Takaaki Hori, and Shinji Watanabe,
\newblock ``Joint {CTC}-attention based end-to-end speech recognition using
  multi-task learning,''
\newblock in {\em Proc. ICASSP}, 2017, pp. 4835--4839.

\bibitem{watanabe2017hybrid}
Shinji Watanabe, Takaaki Hori, Suyoun Kim, John~R Hershey, and Tomoki Hayashi,
\newblock ``{Hybrid CTC/attention architecture for end-to-end speech
  recognition},''
\newblock {\em IEEE Journal of Selected Topics in Signal Processing}, vol. 11,
  no. 8, pp. 1240--1253, 2017.

\bibitem{guo2021recent}
Pengcheng Guo, Florian Boyer, Xuankai Chang, Tomoki Hayashi, Yosuke Higuchi,
  Hirofumi Inaguma, Naoyuki Kamo, Chenda Li, Daniel Garcia-Romero, Jiatong Shi,
  et~al.,
\newblock ``{Recent developments on ESPNET toolkit boosted by conformer},''
\newblock in {\em Proc. ICASSP}, 2021, pp. 5874--5878.

\bibitem{li2020comparison}
Jinyu Li, Yu~Wu, Yashesh Gaur, Chengyi Wang, Rui Zhao, and Shujie Liu,
\newblock ``On the comparison of popular end-to-end models for large scale
  speech recognition,''
\newblock {\em arXiv preprint arXiv:2005.14327}, 2020.

\bibitem{li2020developing}
Jinyu Li, Rui Zhao, Zhong Meng, Yanqing Liu, Wenning Wei, Sarangarajan
  Parthasarathy, Vadim Mazalov, Zhenghao Wang, Lei He, Sheng Zhao, et~al.,
\newblock ``{Developing RNN-T models surpassing high-performance hybrid models
  with customization capability},''
\newblock {\em arXiv preprint arXiv:2007.15188}, 2020.

\bibitem{li2021recent}
Jinyu Li,
\newblock ``Recent advances in end-to-end automatic speech recognition,''
\newblock {\em arXiv preprint arXiv:2111.01690}, 2021.

\bibitem{sainath2018no}
Tara~N Sainath, Rohit Prabhavalkar, Shankar Kumar, Seungji Lee, Anjuli Kannan,
  David Rybach, Vlad Schogol, Patrick Nguyen, Bo~Li, Yonghui Wu, et~al.,
\newblock ``No need for a lexicon? evaluating the value of the pronunciation
  lexica in end-to-end models,''
\newblock in {\em Proc. ICASSP}, 2018, pp. 5859--5863.

\bibitem{Bruguier2016LearningPP}
Antoine Bruguier, Fuchun Peng, and Françoise Beaufays,
\newblock ``Learning personalized pronunciations for contact name
  recognition,''
\newblock in {\em Proc. Interspeech}, 2016.

\bibitem{shenoy_1}
Ashish Shenoy, Sravan Bodapati, Monica Sunkara, Srikanth Ronanki, and Katrin
  Kirchhoff,
\newblock ``Adapting long context {NLM} for {ASR} rescoring in conversational
  agents,''
\newblock in {\em Proc. Interspeech}, 2021.

\bibitem{ptuning}
Saket Dingliwal, Ashish Shenoy, Sravan Bodapati, Ankur Gandhe, Ravi~Teja Gadde,
  and Katrin Kirchhoff,
\newblock ``Domain prompts: Towards memory and compute efficient domain
  adaptation of {ASR} systems,''
\newblock {\em arXiv preprint arXiv:2112.08718}, 2021.

\bibitem{space-efficient}
Christophe Van~Gysel, Mirko Hannemann, Ernest Pusateri, Youssef Oualil, and
  Ilya Oparin,
\newblock ``Space-efficient representation of entity-centric query language
  models,''
\newblock {\em arXiv preprint arXiv:2206.14885}, 2022.

\bibitem{jain2020contextual}
Mahaveer Jain, Gil Keren, Jay Mahadeokar, Geoffrey Zweig, Florian Metze, and
  Yatharth Saraf,
\newblock ``{Contextual RNN-T for open domain ASR},''
\newblock in {\em Proc. Interspeech}, 2020, pp. 11--15.

\bibitem{le2021deep}
Duc Le, Gil Keren, Julian Chan, Jay Mahadeokar, Christian Fuegen, and Michael~L
  Seltzer,
\newblock ``{Deep Shallow Fusion for RNN-T Personalization},''
\newblock in {\em IEEE Spoken Language Technology Workshop (SLT)}, 2021, pp.
  251--257.

\bibitem{gourav2021personalization}
Aditya Gourav, Linda Liu, Ankur Gandhe, Yile Gu, Guitang Lan, Xiangyang Huang,
  Shashank Kalmane, Gautam Tiwari, Denis Filimonov, Ariya Rastrow, et~al.,
\newblock ``{Personalization Strategies for End-to-End Speech Recognition
  Systems},''
\newblock in {\em Proc. ICASSP}, 2021, pp. 7348--7352.

\bibitem{alexadapters}
Kanthashree~Mysore Sathyendra, Thejaswi Muniyappa, Feng-Ju Chang, Jing Liu,
  Jinru Su, Grant~P Strimel, Athanasios Mouchtaris, and Siegfried Kunzmann,
\newblock ``Contextual adapters for personalized speech recognition in neural
  transducers,''
\newblock in {\em Proc. ICASSP}, 2022, pp. 8537--8541.

\bibitem{das2022listen}
Nilaksh Das, Duen~Horng Chau, Monica Sunkara, Sravan Bodapati, Dhanush Bekal,
  and Katrin Kirchhoff,
\newblock ``Listen, {K}now and {S}pell: {K}nowledge-infused subword modeling
  for improving {ASR} performance of {OOV} named entities,''
\newblock in {\em Proc. ICASSP}, 2022, pp. 7887--7891.

\bibitem{clas}
Golan Pundak, Tara~N Sainath, Rohit Prabhavalkar, Anjuli Kannan, and Ding Zhao,
\newblock ``Deep context: end-to-end contextual speech recognition,''
\newblock in {\em IEEE spoken language technology workshop (SLT)}, 2018, pp.
  418--425.

\bibitem{phoebe}
Antoine Bruguier, Rohit Prabhavalkar, Golan Pundak, and Tara~N. Sainath,
\newblock ``Phoebe: Pronunciation-aware contextualization for end-to-end speech
  recognition,''
\newblock in {\em Proc. ICASSP}, 2019, pp. 6171--6175.

\bibitem{shallowfusion}
Anjuli Kannan, Yonghui Wu, Patrick Nguyen, Tara~N Sainath, Zhijeng Chen, and
  Rohit Prabhavalkar,
\newblock ``An analysis of incorporating an external language model into a
  sequence-to-sequence model,''
\newblock in {\em Proc. ICASSP}, 2018, pp. 1--5828.

\bibitem{le2020g2g}
Duc Le, Thilo Koehler, Christian Fuegen, and Michael~L Seltzer,
\newblock ``{G2G: TTS-driven pronunciation learning for graphemic hybrid
  ASR},''
\newblock in {\em Proc. ICASSP}, 2020, pp. 6869--6873.

\bibitem{zhao2021addressing}
Rui Zhao, Jian Xue, Jinyu Li, Wenning Wei, Lei He, and Yifan Gong,
\newblock ``{On Addressing Practical Challenges for RNN-Transducer},''
\newblock {\em arXiv preprint arXiv:2105.00858}, 2021.

\bibitem{attention}
Ashish Vaswani, Noam Shazeer, Niki Parmar, Jakob Uszkoreit, Llion Jones,
  Aidan~N Gomez, {\L}ukasz Kaiser, and Illia Polosukhin,
\newblock ``Attention is all you need,''
\newblock {\em Advances in neural information processing systems}, vol. 30,
  2017.

\bibitem{liptchinsky2017letter}
Vitaliy Liptchinsky, Gabriel Synnaeve, and Ronan Collobert,
\newblock ``Letter-based speech recognition with gated convnets,''
\newblock {\em arXiv preprint arXiv:1712.09444}, 2017.

\bibitem{voxpopuli}
Changhan Wang, Morgane Riviere, Ann Lee, Anne Wu, Chaitanya Talnikar, Daniel
  Haziza, Mary Williamson, Juan Pino, and Emmanuel Dupoux,
\newblock ``Voxpopuli: A large-scale multilingual speech corpus for
  representation learning, semi-supervised learning and interpretation,''
\newblock {\em arXiv preprint arXiv:2101.00390}, 2021.

\bibitem{watanabe2018espnet}
Shinji Watanabe, Takaaki Hori, Shigeki Karita, Tomoki Hayashi, Jiro Nishitoba,
  Yuya Unno, Nelson {Enrique Yalta Soplin}, Jahn Heymann, Matthew Wiesner,
  Nanxin Chen, Adithya Renduchintala, and Tsubasa Ochiai,
\newblock ``{ESPnet}: End-to-end speech processing toolkit,''
\newblock in {\em Proc. Interspeech}, 2018, pp. 2207--2211.

\end{thebibliography}

\end{document}